\documentclass[conference]{IEEEtran}
\usepackage{booktabs}
\IEEEoverridecommandlockouts
\usepackage{marvosym}
\usepackage{cite}
\usepackage{amsmath,amssymb,amsfonts}
\usepackage{algorithmic}
\usepackage{graphicx}
\usepackage{textcomp}
\usepackage{xcolor}
\usepackage{multirow}
\def\BibTeX{{\rm B\kern-.05em{\sc i\kern-.025em b}\kern-.08em
    T\kern-.1667em\lower.7ex\hbox{E}\kern-.125emX}}
\begin{document}

\title{Linear Multi-granular Body Modeling for Text-Driven Motion Generation}

\author{
Xingzu Zhan\textsuperscript{1,\thanks{* Equal contribution. \textsuperscript{\Letter} Corresponding Author.}*} \quad
Genpei Zhang\textsuperscript{2,*} \quad
Chen Xie\textsuperscript{1} \quad
Honghang Chen\textsuperscript{1} \quad 
Haoran Sun\textsuperscript{1} \quad
Xiaochun Mai\textsuperscript{1,\Letter}\\[5pt]
\textsuperscript{1}~Shenzhen University \quad
\textsuperscript{2}~University of Electronic Science and Technology of China
\thanks{Xiaochun Mai is with the College of Electronics and Information Engineering, Shenzhen University, Shenzhen, China, with the Shenzhen Key Laboratory of Digital Creative Technology, with the Guangdong Province Engineering Laboratory for Digital Creative Technology, and also with the Guangdong-Hong Kong Joint Laboratory for Big Data Imaging and Communication, Shenzhen, China.}
}

\maketitle

\begin{abstract}
Text-to-motion generation is a rapidly growing field at the nexus of multimodal learning and computer graphics, enabling flexible and cost-effective applications in gaming, animation, robotics, and virtual reality. 
Existing part-aware pipelines typically split the body into segments and then recover whole-body coordination via attention-based fusion, which incurs quadratic complexity and fails to model part-level and whole-body cues simultaneously. 
Moreover, we observe that naively stacking spatial and temporal modules accumulates redundant features, thereby requiring deeper networks to achieve competitive fidelity. 
To address these issues, we propose HiSTF Mamba, a linear-time framework composed of \emph{Dual-Spatial Mamba}, \emph{Bi-Temporal Mamba}, and the \emph{Dynamic Spatiotemporal Fusion Module} (DSFM). 
Dual-Spatial Mamba performs parallel part-based and whole-based modeling to simultaneously capture local joint dynamics and global coordination. 
Bi-Temporal Mamba employs bidirectional scanning to encode both short-term motion details and long-term dependencies. 
DSFM further removes redundancy and extracts complementary temporal information before fusing it with spatial features, yielding an expressive spatiotemporal representation. 
Extensive experiments on HumanML3D and KIT-ML demonstrate that HiSTF Mamba improves generation fidelity and semantic alignment while requiring substantially fewer parameters than existing segment-based architectures.
\end{abstract}


\section{Introduction}
\label{sec:intro}

Text-to-motion aims to generate human motion sequences based on natural language descriptions, and has emerged as a key research direction at the intersection of multimodal learning and computer graphics. Using textual input to drive human motion generation, text-to-motion provides a flexible and cost-effective approach for applications such as games~\cite{Lee2002}, animation~\cite{Kappel2021}, robotic interaction~\cite{Koppula2015}, and virtual reality (VR).

Text-to-motion generation has evolved from GAN and VAE models~\cite{Xu2023,Bie2022,Komura2017,Ghosh2021} to diffusion frameworks~\cite{tevet2023human,chen2023executing,zhang2022motiondiffuse}, eventually incorporating architectures that explicitly decouple spatial and temporal modeling~\cite{zhang2023finemogen,liu2023spatio,zhang2024motion}. While Motion Mamba~\cite{zhang2024motion} and T2M Mamba~\cite{11696326}replaces transformers with State Space Models to achieve efficient long sequence modeling, it continues to process the human skeleton as a unified entity. This holistic approach often neglects subtle joint level dynamics and relies on stacked spatiotemporal blocks, leading to feature redundancy. To capture these intricate limb movements, recent studies have shifted toward segment-based architectures. Methods such as SCA~\cite{Ghosh2021}, AttT2M~\cite{zhong2023attt2m}, and ParCo~\cite{zou2024parco} typically adopt a strategy where body parts are encoded independently and subsequently coupled via attention mechanisms to restore global coordination. However, this independent encoding followed by aggregation frequently incurs high computational costs due to the quadratic complexity of attention. Furthermore, by enforcing cross-part coupling only at the final aggregation stage, these methods rely on late-stage correction to recover global coherence, which is prone to unstable synchronization and poor inter-limb coherence, especially for long sequences.

\begin{figure}[t]
  \centering
  \includegraphics[width=0.5\textwidth]{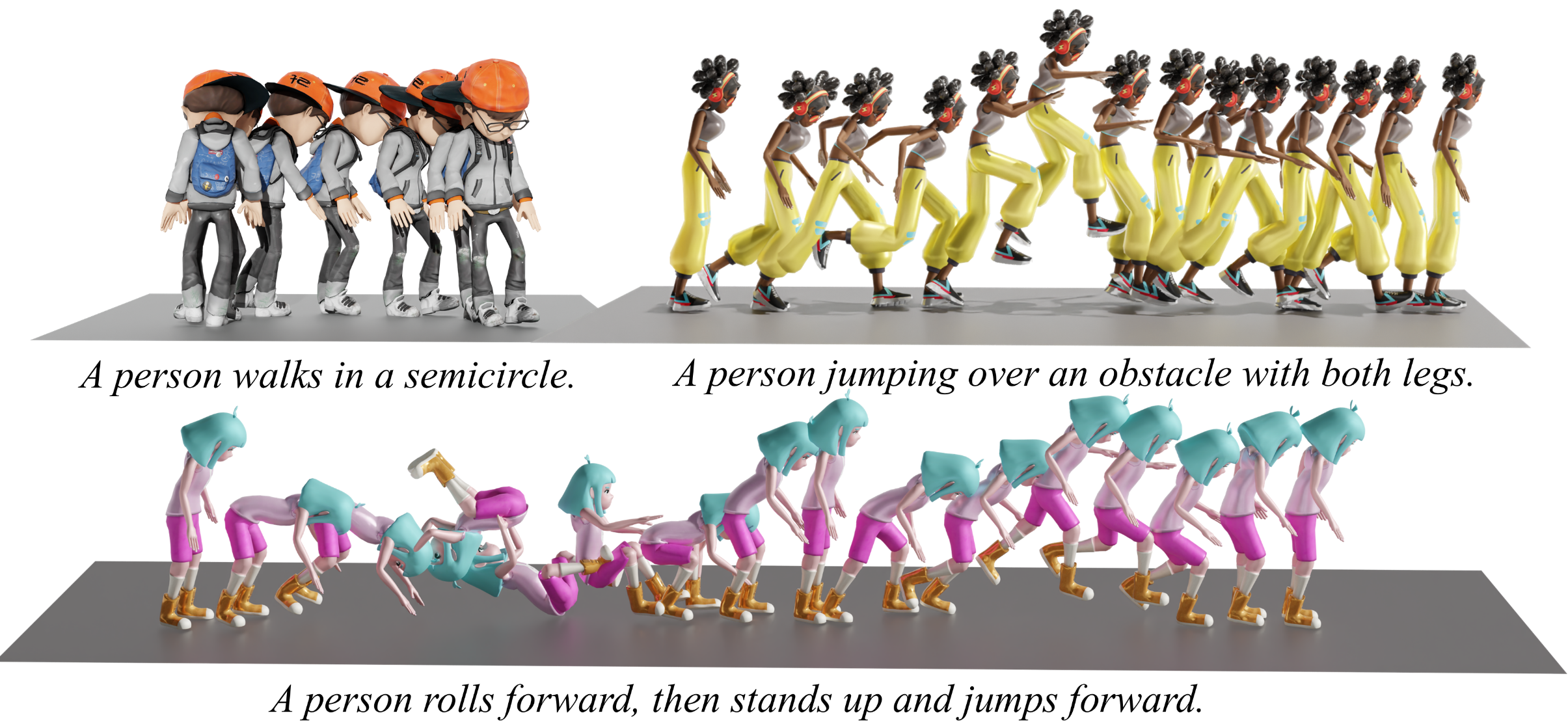}
  \vspace{-7mm}
  \caption{Qualitative visualization of motions generated by HiSTF Mamba.}
  \label{fig:fig1}
  \vspace{-4mm}   
\end{figure}

To alleviate the high computational cost of attention-based part-aware pipelines and the poor inter-limb coherence caused by enforcing cross-part coupling only at the final fusion stage, we propose HiSTF Mamba. First, Dual-Spatial Mamba models fine joint dynamics and whole-body coordination in parallel via a "Part-based + Whole-based" strategy, avoiding expensive attention. Bi-Temporal Mamba adopts bidirectional selective SSMs to capture short-range cues and long-range dependencies for temporally coherent synthesis. We further observe that naively stacking decoupled spatial and temporal modules accumulates redundant features, which often forces deeper and heavier networks. To mitigate this, we introduce the Dynamic Spatiotemporal Fusion Module (DSFM), which filters redundant temporal signals, extracts complementary information across temporal features, and then fuses the refined temporal representation with spatial outputs. This hierarchical fusion yields a compact yet expressive spatiotemporal representation, leading to more coordinated full-body motions and sharper joint-level details, as illustrated in Fig.~\ref{fig:fig1}. 

Comprehensive evaluations on the HumanML3D and KIT-ML datasets substantiate the robust capabilities of HiSTF Mamba across multiple metrics. A distinguishing feature of our framework is its efficiency, compared to ParCo~\cite{zou2024parco}, which employs computationally intensive attention mechanisms and necessitates 168.4M parameters, HiSTF Mamba utilizes only 23.95M parameters. Despite this substantial reduction in complexity, our method secures superior performance in terms of generation fidelity and semantic alignment.

Overall, our contributions to the text-to-motion community can be summarized as follows:

\begin{enumerate}
    \item We propose the Dual-Spatial Mamba module, which utilizes a "Part-based + Whole-based" parallel strategy to capture both whole-body coordination and fine-grained joint-level details simultaneously, achieving robust spatial modeling with linear computational complexity.

    \item  We design a Dynamic Spatiotemporal Fusion Module (DSFM) to perform redundancy removal and extraction of complementary information for temporal features, which is then fused with spatial features to construct a more efficient spatio-temporal representation.

    \item Comprehensive evaluations on the HumanML3D and KIT-ML benchmarks validate the effectiveness of our approach. Compared to ParCo, our model reduces the parameter count from 168.4M to 23.95M and improves the FID score by approximately 64.2\% as it decreases from 0.109 to 0.039, demonstrating a superior balance between generation quality and model efficiency.
\end{enumerate}

\vspace{-1mm}
\section{Method}
\label{method}

This section details the HiSTF Mamba architecture for text driven human motion generation. We first present the overall framework, followed by the core spatiotemporal components: Dual-Spatial Mamba and Bi-Temporal Mamba. Finally, we describe the Dynamic Spatiotemporal Fusion Module (DSFM).

\label{sec:Spatiotemporal Motion Mamba}

\subsection{HiSTF Mamba}

\begin{figure*}[t]
  \centering
  \includegraphics[width=0.93\textwidth,
                   height=0.40\textheight,
                   keepaspectratio]{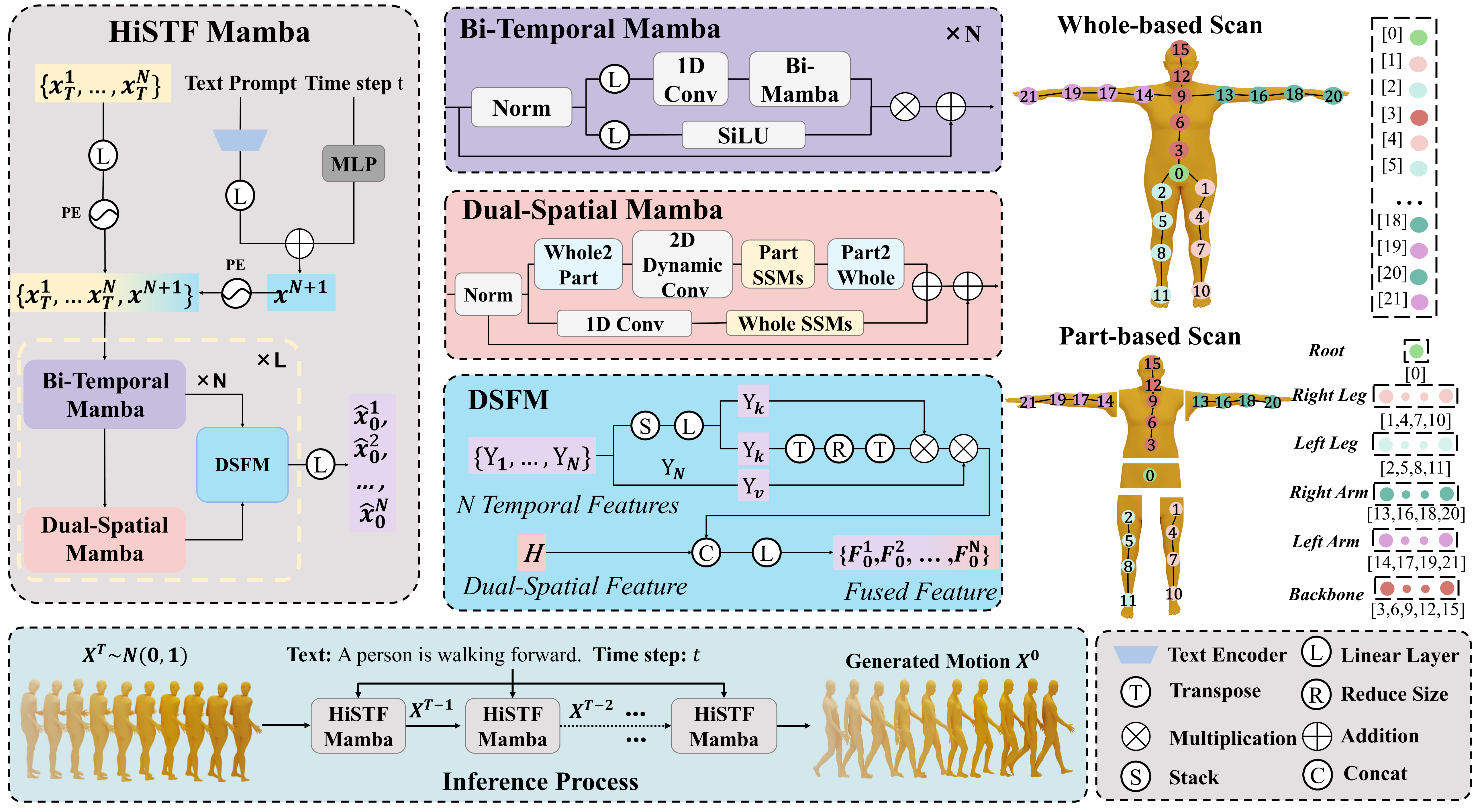}
  \vspace{-2mm}
    \caption{\textbf{The overview of the proposed HiSTF Mamba.} Given a text prompt and diffusion timestep, the model iteratively denoises from Gaussian noise to synthesize a motion sequence, as illustrated by the inference process at the bottom.
  The backbone consists of N \emph{Bi-Temporal Mamba} that encodes bidirectional temporal features $\{T_1,\ldots,T_N\}$,
  a \emph{Dual-Spatial Mamba} that captures multi-granular spatial cues via parallel \emph{part-based} and \emph{whole-based} scans,
  and a \emph{DSFM} that suppresses redundant temporal information and fuses refined temporal cues with spatial features.
  The right panel visualizes scan order of whole SSMs and the Part SSMs.}
    \label{fig:fig2}
\vspace{-5mm}
\end{figure*}

To achieve linear complexity while simultaneously modeling whole body coordination and fine grained limb details, we propose \textbf{HiSTF Mamba}, a hierarchical spatiotemporal fusion approach, as illustrated in Fig.~\ref{fig:fig2}. Through three specialized components: \emph{N} Bi-Temporal Mamba blocks, a Dual-Spatial Mamba block, and a DSFM module, HiSTF Mamba effectively captures fine-grained joint details and fuses spatiotemporal features. Specifically, the input sequence is first processed by \emph{N} layers of Bi-Temporal Mamba to model rich temporal dependencies, and the resulting features are then fed into the Dual-Spatial Mamba block. Within this block, 2D dynamic convolution and Part SSMs extract joint-detail features from six body segments, which are then combined with the Whole SSMs’ global coordination to obtain the spatial feature $S$. Next, the \emph{N} temporal features $\{T_{1},\ldots,T_{N}\}$ are passed to the DSFM for redundancy removal and subsequently fused with $S$, yielding a spatiotemporal motion representation that preserves both joint-level detail and overall body coordination.

In the sampling stage, we first encode the text condition with the
CLIP~\cite{radford2021learning} text encoder, apply random masking, and
project it into the latent space.  Meanwhile, we project the noise step
$T$ through an MLP and add it to the encoded text condition, obtaining
$x^{(N+1)}$.  we incorporate the $x^{(N+1)}$ with the motion sequence subjected to $T$ steps of noise,
yielding
$\{x_T^1, x_T^2, \ldots, x_T^N\}$, which is fed into the model to
predict the clean sequence  $\{\hat{{x}}_{0}^{1},\hat{{x}}_{0}^{2},\dots,\hat{{x}}_{0}^{N}\}$.  The diffusion process then
generates $x^{T-1}$ from  $\{\hat{{x}}_{0}^{1},\hat{{x}}_{0}^{2},\dots,\hat{{x}}_{0}^{N}\}$.  Repeating this procedure $T$
times ultimately produces a clean sequence  $\{{{x}}_{0}^{1},{{x}}_{0}^{2},\dots,{{x}}_{0}^{N}\}$ that conforms to the
text condition.  This design balances global motion patterns with
fine-grained joint dynamics, producing more accurate reconstructions. Unipc scheduler~\cite{zhao2023unipc} is used to accelerate sampling.

\vspace{-0.5mm}
\subsection{Bi-Temporal Mamba and Dual-Spatial Mamba}
As shown in Figure~\ref{fig:fig2}, 
Bi-Temporal Mamba employs a bidirectional scanning mechanism to capture the contextual information of motion sequences in both forward and backward directions, thereby generating more coordinated and coherent motion sequences. 
Dual-Spatial Mamba adopts parallel Part-based and Whole-based learning, effectively 
capturing motion features that exhibit both fine-grained joint details and whole-body coordination.

\textbf{Bi-Temporal Mamba.} To generate coherent and smooth motion sequences, we employ Selected SSMs featuring a bidirectional scanning mechanism. We begin with an input sequence \(T_{l-1} \in \mathbb{R}^{(B \times T \times D)}\), where \(B\) is the batch size, \(T\) is the sequence length, and \(D\) is the feature dimension. First, \(T_{l-1}\) is normalized and subsequently projected by a linear layer into a new representation space \(\mathbb{R}^{(B \times T \times E)}\), yielding two transformed sequences \(X\) and \(Z\). We then perform forward and backward scans over \(X\), each governed by learnable parameters \(\Delta\), \(B\), and \(C\), which dynamically adjust the state update equation.
\begin{equation}
\begin{aligned}
\begin{cases}
h_{t} = A(\Delta_t)\,h_{t-1} + B_t\,x_t,\\[6pt]
y_{t} = C_t\,h_{t},
\end{cases}
\quad
\begin{cases}
B_t &= s_B\bigl(x_t\bigr),\\
C_t &= s_C\bigl(x_t\bigr),\\
\Delta_t &= \tau_\Delta \!\bigl(W_\Delta \, x_t\bigr).
\end{cases}
\end{aligned}
\label{eq:two}
\end{equation}

The forward and backward outputs are then elementwise multiplied with the nonlinear activation \(\mathrm{SiLU}(Z)\) and fused. Finally, a linear projection maps the fused representation back to the original dimension \(\mathbb{R}^{(B \times T \times D)}\).This bidirectional scanning strategy is particularly effective for modeling motion sequences. The forward branch captures how earlier actions influence and shape the current state, while the backward branch leverages future actions as contextual cues. By combining these two directions within a temporal recursion framework, the model more thoroughly encodes transitions and dependencies across time steps, ultimately producing more coordinated and smooth motion sequences.

\textbf{Dual-Spatial Mamba.} To simultaneously capture whole body coordination and intricate joint dynamics with linear complexity, we propose the Dual-Spatial Mamba. This module performs two types of learning at the same time through Multi-Granular Body-Spatial Modelling, as described in Figure~\ref{fig:fig2}: on the Part-based the entire motion matrix is decomposed into 6 body part matrices, so that each input is limited to body part joints, allowing the model to focus on learning body part joint details, avoiding too much joint information being input at the same time, which can be severely compressed early on. The Whole-based retains the entire motion matrix inputs to learn whole-body motion coordination. Thus, Dual-Spatial Mamba adopts Part-based and Whole-based parallel learning to capture both local and global motion features, which are then fused via an element-wise addition strategy.

Specifically, the input sequence is normalized. In the Part-based learning phase, we first map the normalized full-body motion sequence back to its original dimensionality. Drawing upon insights from ParCo~\cite{zou2024parco}, as shown in Fig~\ref{fig:fig2}, we design a Whole2Part module to split the whole-body motion into six parts. Right Leg, Left Leg, Backbone, Root, Right Arm, and Left Arm. This design follows 
the common practice of 3D human body models (e.g., SMPL, MMM), where Kinematic Trees typically represent the skeleton as five main chains (the four limbs plus the backbone); We adds a separate Root component to account for global trajectories, thereby leading to six sub-motions in total. Formally, given a motion sequence $x = \{x^1, x^2, x^3, \dots, x^N\}$, it is partitioned into $p_i = \{p_i^1, p_i^2, p_i^3, \dots, p_i^N\}, \quad i \in \{1, 2, \ldots, S\}$, where \(S\) denotes the total number of body parts.
Upon partitioning the motion, each of the six anatomical segments is projected and processed via 2D dynamic convolutions using six learnable kernels to extract specific patterns. These component features are stacked along the batch dimension and fed into the Part SSMs to model local dependencies, after which the Part2Whole operation restores the holistic body structure. In parallel, the global branch employs 1D convolutions and Whole SSMs to capture long range inter joint relationships. Finally, we integrate the local and global features via element wise summation to yield a comprehensive spatiotemporal representation.


\label{experiments}
\begin{table*}[ht]
    \centering
    \small
    \caption{Comparison of results on the HumanML3D and KIT-ML test sets. The right arrow ($\rightarrow$) indicates that values closer to real motions correspond to better performance. Each evaluation was run 20 times, and ``$\pm$'' denotes the 95\% confidence interval. Bold text highlights the best scores, while underlined entries mark the second best.}
    \vspace{-1.5mm}
    \resizebox{0.96\textwidth}{!}{
    \begin{tabular}{l@{\hspace{0.3cm}}c@{\hspace{0.4cm}}c@{\hspace{0.4cm}}c@{\hspace{0.4cm}}c@{\hspace{0.4cm}}c@{\hspace{0.4cm}}c@{\hspace{0.4cm}}c@{\hspace{0.4cm}}c}
        \toprule
        
        \multirow{2}{*}{\hspace{12mm} Methods} & \multicolumn{3}{c}{R Precision$\uparrow$} & \multirow{2}{*}{FID$\downarrow$} & \multirow{2}{*}{MM Dist$\downarrow$} & \multirow{2}{*}{Diversity$\rightarrow$} & \multirow{2}{*}{MModality$\uparrow$} & \multirow{2}{*}{Params (M)$\downarrow$} \\
        \cmidrule(lr){2-4}
        & Top 1 & Top 2 & Top 3 & & & & & \\
        \midrule
        \hspace{7mm}\textit{On the HumanML3D} & 
        \textit{0.511}$^{\pm .003}$ & 
        \textit{0.703}$^{\pm .003}$ & 
        \textit{0.797}$^{\pm .002}$ & 
        \textit{0.002}$^{\pm .000}$ & 
        \textit{2.974}$^{\pm .008}$ & 
        \textit{9.503}$^{\pm .065}$ & -- & -- \\

        \midrule
        T2M~\cite{guo2022generating}  & 0.457$^{\pm .002}$ & 0.639$^{\pm .003}$ & 0.740$^{\pm .003}$ & 1.067$^{\pm .002}$ & 3.340$^{\pm .008}$ & 9.188$^{\pm .002}$ & 2.090$^{\pm .083}$ & 31.16 \\
        MDM~\cite{tevet2023human} & 0.320$^{\pm .005}$ & 0.498$^{\pm .004}$ & 0.611$^{\pm .007}$ & 0.544$^{\pm .044}$ & 5.566$^{\pm .027}$ & \textbf{9.559}$^{\pm .086}$ & \textbf{2.799}$^{\pm .072}$ & \textbf{17.88} \\
        MotionDiffuse~\cite{zhang2022motiondiffuse} & 0.491$^{\pm .001}$ & 0.681$^{\pm .001}$ & 0.782$^{\pm .001}$ & 0.630$^{\pm .001}$ & 3.113$^{\pm .001}$ & 9.410$^{\pm .049}$ & 1.553$^{\pm .042}$ & 87.15 \\
        MLD~\cite{chen2023executing} & 0.481$^{\pm .003}$ & 0.673$^{\pm .003}$ & 0.772$^{\pm .002}$ & 0.473$^{\pm .013}$ & 3.196$^{\pm .010}$ & 9.724$^{\pm .082}$ & 2.413$^{\pm .079}$ & 26.38 \\
        Motion Mamba~\cite{zhang2024motion} & 0.502$^{\pm .003}$ & 0.693$^{\pm .002}$ & 0.792$^{\pm .002}$ & 0.281$^{\pm .009}$ & 3.060$^{\pm .058}$ & 9.871$^{\pm .084}$ & 2.294$^{\pm .058}$ & -- \\
        T2M-GPT~\cite{zhang2023t2mgptgeneratinghumanmotion} & 0.492$^{\pm .003}$ & 0.679$^{\pm .002}$ & 0.775$^{\pm .002}$ & 0.141$^{\pm .005}$ & 3.121$^{\pm .009}$ & 9.722$^{\pm .082}$ & 1.831$^{\pm .048}$ & 247.80 \\
        AttT2M~\cite{zhong2023attt2m} & 0.499$^{\pm .003}$ & 0.690$^{\pm .002}$ & 0.786$^{\pm .002}$ & 0.112$^{\pm .006}$ & 3.038$^{\pm .007}$ & 9.700$^{\pm .090}$ & \underline{2.452}$^{\pm .051}$ & -- \\
        ParCo~\cite{zou2024parco} & 0.515$^{\pm .003}$ & 0.706$^{\pm .003}$ & 0.801$^{\pm .002}$ & 0.109$^{\pm .005}$ & \underline{2.927}$^{\pm .008}$ & \underline{9.576}$^{\pm .088}$ & 1.382$^{\pm .060}$ & 168.4 \\
        MoMask~\cite{guo2024momask} & \underline{0.521}$^{\pm .002}$ & \underline{0.713}$^{\pm .002}$ & \underline{0.807}$^{\pm .002}$ &  \underline{0.045}$^{\pm .002}$ & 2.958$^{\pm .008}$ & - & 1.241$^{\pm .040}$ & 44.85 \\ 
        \midrule
        HiSTF Mamba (Ours) & \textbf{0.537}$^{\pm .002}$ & \textbf{0.724}$^{\pm .002}$ & \textbf{0.813}$^{\pm .002}$ & \textbf{0.039}$^{\pm .002}$ & \textbf{2.884}$^{\pm .008}$ & 9.689$^{\pm .091}$ &  1.701$^{\pm .064}$ & \underline{23.95} \\
        
        \midrule
        \hspace{9mm}\textit{On the KIT-ML} & 
        \textit{0.424}$^{\pm .005}$ & 
        \textit{0.649}$^{\pm .006}$ & 
        \textit{0.779}$^{\pm .006}$ & 
        \textit{0.031}$^{\pm .004}$ & 
        \textit{2.788}$^{\pm .012}$ & 
        \textit{11.08}$^{\pm .097}$ & -- & -- \\
        \midrule
        T2M ~\cite{guo2022generating} & 0.370$^{\pm .005}$ & 0.569$^{\pm .007}$ & 0.693$^{\pm .007}$ & 2.770$^{\pm .109}$ & 3.401$^{\pm .008}$ & 10.91$^{\pm .119}$ & 1.482$^{\pm .065}$ & -- \\
        MDM ~\cite{tevet2023human} & 0.164$^{\pm .004}$ & 0.291$^{\pm .004}$ & 0.396$^{\pm .004}$ & 0.497$^{\pm .021}$ & 9.191$^{\pm .022}$ & 10.85$^{\pm .109}$ & 1.907$^{\pm .214}$ & -- \\
        MotionDiffuse ~\cite{zhang2022motiondiffuse} & 0.417$^{\pm .004}$ & 0.621$^{\pm .004}$ & 0.739$^{\pm .004}$ & 1.954$^{\pm .062}$ & 2.958$^{\pm .005}$ & \textbf{11.10}$^{\pm .143}$ & 0.730$^{\pm .013}$ & -- \\
        MLD ~\cite{chen2023executing} & 0.390$^{\pm .008}$ & 0.609$^{\pm .008}$ & 0.734$^{\pm .007}$ & 0.404$^{\pm .027}$ & 3.204$^{\pm .027}$ & 10.80$^{\pm .117}$ & \underline{2.192}$^{\pm .071}$ & -- \\
        Motion Mamba~\cite{zhang2024motion} & 0.419$^{\pm .006}$ & 0.645$^{\pm .005}$ & 0.765$^{\pm .006}$ & 0.307$^{\pm .041}$ & 3.021$^{\pm .025}$ & \underline{11.02}$^{\pm .098}$ & 1.678$^{\pm .064}$ & -- \\
        T2M-GPT~\cite{zhang2023t2mgptgeneratinghumanmotion} & 0.416$^{\pm .006}$ & 0.627$^{\pm .006}$ & 0.745$^{\pm .006}$ & 0.514$^{\pm .029}$ & 3.007$^{\pm .023}$ & 10.92$^{\pm .108}$ & 1.570$^{\pm .039}$ & -- \\
        AttT2M~\cite{zhong2023attt2m} & 0.413$^{\pm .006}$ & 0.632$^{\pm .006}$ & 0.751$^{\pm .006}$ & 0.870$^{\pm .039}$ & 3.039$^{\pm .021}$ & 10.96$^{\pm .123}$ & \textbf{2.281}$^{\pm .047}$ & -- \\
        ParCo~\cite{zou2024parco} & 0.430$^{\pm .004}$ & 0.649$^{\pm .007}$ & 0.772$^{\pm .006}$ & 0.453$^{\pm .027}$ & 2.820$^{\pm .028}$ & 10.95$^{\pm .094}$ & 1.245$^{\pm .022}$ & -- \\
        MoMask~\cite{guo2024momask} & \underline{0.433}$^{\pm .007}$ & \underline{0.656}$^{\pm .005}$ & \underline{0.781}$^{\pm .005}$ & \textbf{0.204}$^{\pm .011}$ & \underline{2.779}$^{\pm .022}$ & - & 1.131$^{\pm .043}$ & -- \\  
        \midrule
        HiSTF Mamba (Ours) & \textbf{0.458}$^{\pm .006}$ & \textbf{0.678}$^{\pm .007}$ & \textbf{0.798}$^{\pm .004}$ & \underline{0.217}$^{\pm .009}$ & \textbf{2.737}$^{\pm .017}$ & 11.16$^{\pm .217}$ &  1.058$^{\pm .028}$ & -- \\
        
        \bottomrule
    \end{tabular}}
    \label{tab:humanml3d}
    \vspace{-4mm}
\end{table*}

\vspace{-1mm}
\subsection{Dynamic Spatiotemporal Fusion Module}
We observe that the simple stacking of spatial and temporal modules induces feature redundancy, which typically necessitates deep and parameter heavy networks to capture effective patterns. To mitigate this and ensure model efficiency, inspired by the DMCA module in Sparx~\cite{Lou2024},  we propose a dynamic spatiotemporal fusion module (DSFM). Differently, DSFM not only reduces redundancy and extracts complementary features among the output features from the \( N \) layers of Temporal Mamba, but also modifies the fusion strategy to effectively integrate this refined temporal features with spacial features to obtain a expressive spatiotemporal representation.
Specifically, as shown in Fig.~\ref{fig:fig2}, let \( Y \in \mathbb{R}^{(C \times T \times D)} \) represent the output features of \( N \) Temporal Mamba layers, where \( Y = \{ Y_1, Y_2, \dots, Y_N \} \).
In practice, the output of Temporal Mamba is originally in the form \( Y \in \mathbb{R}^{(T \times D)} \).
To facilitate an effective division into \( Y_q \) and \( Y_k \) in the subsequent processing steps, we apply an unsqueeze operation to reshape it into \( Y \in \mathbb{R}^{(C \times T \times D)} \).
Additionally, let \( H \in \mathbb{R}^{(T \times D)} \) denote the output features of Dual-Spatial Mamba.
Here, \( C \) represents the number of channels, while \( T \) and \( D \) denote the sequence length and the feature dimension respectively.

To obtain refined temporal information, the first $N-1$ timporal features \(\{Y_1, Y_2, \dots, Y_{N-1}\}\) are stacked along the channel dimension and then projected into \(\mathbb{R}^{(2C \times  T \times D)}\). The projected features are subsequently split into two components, each squeezed to \(\mathbb{R}^{(T \times D)}\). One component serves as the query \(Y_q\), while the other serves as the key \(Y_k\).

The attention matrix is computed using \(Y_v\) and \(Y_k\). According to \cite{Lou2024}, this attention matrix is independent of the spatial dimension of \(Y_v\) and \(Y_k\). Hence, a reduction by a factor of \(s\) can be applied, mapping \(Y_k, Y_v \) to \(\mathbb{R}^{(T \times \tfrac{D}{s})}\). This step considerably lowers computational overhead and memory usage while preserving key temporal relationships. The resulting matrix is then used to weight the query \(Y_q\), yielding the refined temporal feature.

Next, the refined temporal features are integrated with the spatial features.
The resulting representations are concatenated along the feature dimension,
forming \(\mathbb{R}^{(T \times 2D)}\). Finally, a linear projection is applied to map the concatenated features back to
\(\mathbb{R}^{(T \times D)}\), ensuring a unified representation.
This fusion mechanism allows the model to effectively capture complementary
spatiotemporal patterns, enhancing its ability to learn robust feature
representations.

Mathematically, the Dynamic Spatiotemporal Fusion Module (DSFM) can be formulated as follows:
\begin{equation}
\begin{aligned}
Y_k, Y_q &= \text{squeeze}\!\Bigl(\text{split}\bigl(W_1 \cdot \text{concat}(Y_1, Y_2, \dots, Y_{N-1})\bigr)\Bigr), \\
V &= R\bigl((Y_N)^T\bigr), \quad K = R\bigl((Y_k)^T\bigr), \quad Q = Y_q, \\
\mathbf{T} &= \frac{\text{Softmax}(QK^T)}{\sqrt{N/r}} \, V, \quad Z = W_2 \cdot \text{Concat}\bigl( S, T\bigr).
\end{aligned}
\label{eq:mygroup}
\end{equation}
where \( W \) represents a linear projection. Specifically, \( W_1 \in \mathbb{R}^{((N-1)C \times 2C)} \) performs a projection along the channel dimension, while \( W_2 \in \mathbb{R}^{(2D \times D)} \) projects along the feature dimension. The function \( R \) denotes a reduction operation that decreases the feature dimension, implemented via a strided convolution. Additionally, \( K, V \in \mathbb{R}^{(T \times \tfrac{D}{s})} \) represent the key and value matrices after dimensionality reduction.

\section{Experiments}
\begin{figure*}[t]
  \centering
  \includegraphics[width=0.95\textwidth]{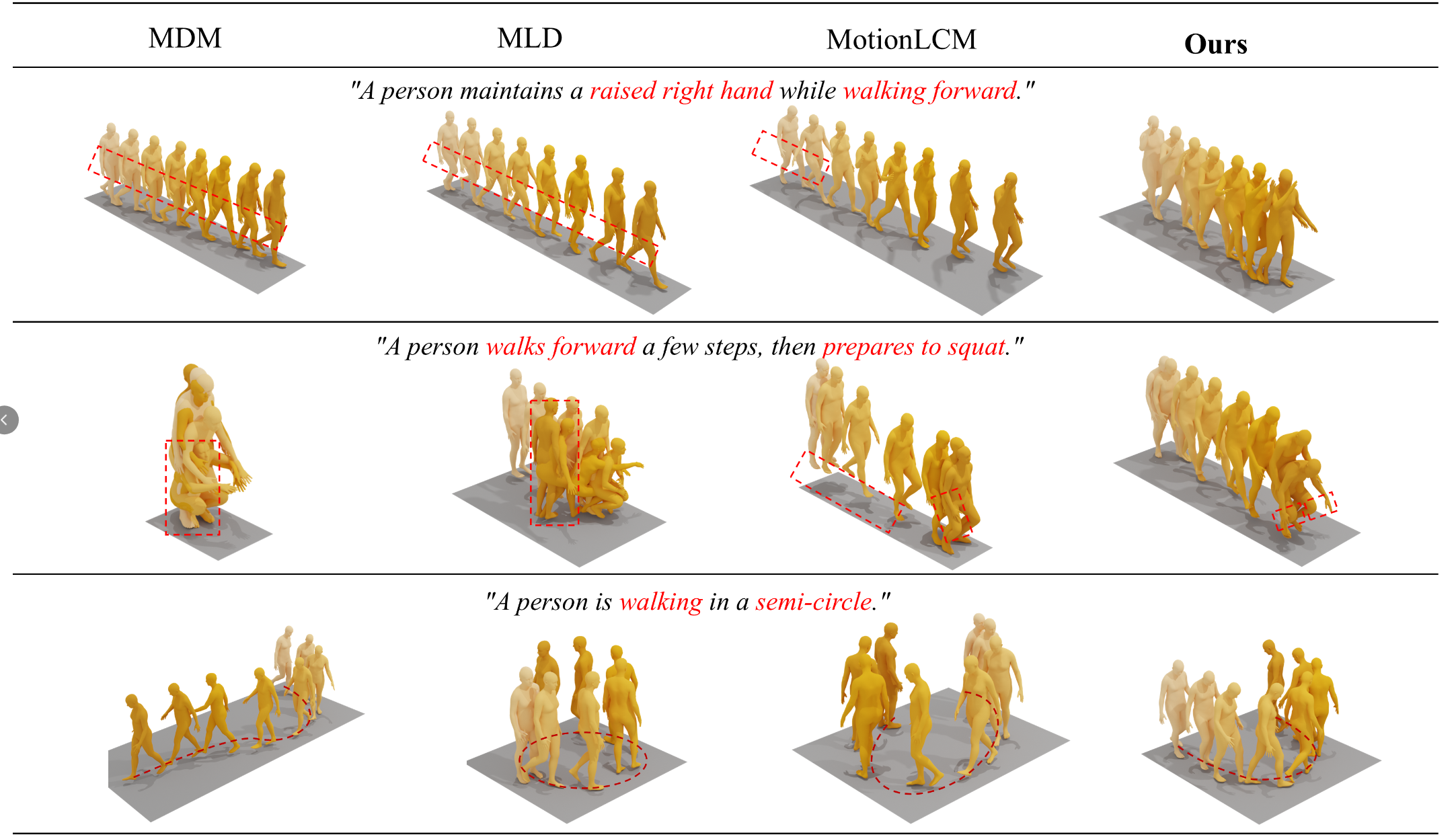}%
  
  \caption{ \textbf{Qualitative comparison of typical methods}. We randomly selected three textual samples to generate motion sequences. Our results exhibit closer alignment with the text and present richer joint-level details.}
 \vspace{-1mm}
  \label{fig:fig3}
\end{figure*}

\subsection{Datasets}
To evaluate the effectiveness of our method, we conduct experiments on two widely-used text-to-motion datasets: HumanML3D~\cite{guo2022generating} and KIT-ML~\cite{plappert2016kit}. HumanML3D consists of 14,616 motion sequences and 44,970 textual descriptions. Meanwhile, KIT-ML, introduced by Plappert et al., comprises 3,911 motion sequences and 6,278 natural language annotations. To ensure a fair comparison, we follow the evaluation protocol established in \cite{guo2022generating} for subsequent experiments.

\subsection{Comparative Studies}
We conducted evaluations on the HumanML3D~\cite{guo2022generating} and KIT-ML~\cite{plappert2016kit} datasets to benchmark our method against several leading approaches. By adopting the network structure detailed in Section~\ref{sec:Spatiotemporal Motion Mamba} with $N=3$ and $L=2$, our method demonstrates robust performance improvements as reported in Table~\ref{tab:humanml3d}. On the HumanML3D dataset, our model achieves superior semantic alignment and secures the highest R-Precision scores alongside the lowest Multimodal Distance. Notably, our approach attains a state-of-the-art FID score of 0.039, surpassing the 0.045 score of MoMask. Furthermore, our framework exhibits significant efficiency advantages by utilizing only 23.95M parameters, which represents nearly half the model size required by MoMask. Regarding the KIT-ML dataset, our method similarly excels in text consistency by achieving the best R-Precision across all ranks and the lowest MM Distance of 2.737, while maintaining competitive fidelity with an FID of 0.217. Additionally, the qualitative results in Fig.~\ref{fig:fig3} illustrate that our generated motions exhibit closer semantic alignment with textual descriptions, finer joint dynamics, and superior whole-body coordination compared to other approaches.

\subsection{Ablation Studies}
In this experiment, we investigate how the number of Bi-Temporal Mamba blocks ($N$), the total number of layers ($L$), and the presence or absence of key components affect the model's performance on the HumanML3D dataset.

 \textbf{Ablation of layer.} As presented in Table~\ref{tab:Layers}, we analyze how varying the quantity of Bi-Temporal Mamba blocks ($N$) and the total layers ($L$) affects model efficacy. With the configuration of $N=3$ and $L=2$, our method attains the minimum Fréchet Inception Distance (FID) of $0.039$, signifying a superior capture of the intrinsic motion distribution. While increasing $N$ to 4 yields marginal gains in R Precision, it results in a degradation of FID and raises computational demands. Consequently, we identify $N=3$ and $L=2$ as the optimal choice to maintain a balance between generation quality and efficiency.

\textbf{Ablation of key components.} Table~\ref{core module} presents the ablation results regarding the contribution of each module. We first investigate the impact of the Dynamic Spatiotemporal Fusion Module (DSFM). The results show that removing this component causes the FID score to rise significantly from 0.039 to 0.096. This substantial degradation demonstrates that the DSFM is crucial for effectively fusing spatial and temporal features to ensure motion realism. Subsequently, we examine the Dual Spatial Mamba by isolating its Part based and Whole based branches. The data indicates that the full model achieves superior fidelity compared to configurations lacking either branch. Specifically, excluding the Part based learning increases the FID to 0.066, while removing the Whole based learning yields an FID of 0.065. Furthermore, the simultaneous removal of both branches results in an FID of 0.070. Although the configuration without the Part based branch exhibits a marginally lower Multi Modal Distance, the comprehensive HiSTF Mamba framework secures the lowest FID score overall. These findings confirm that integrating global body coordination with fine grained joint details is essential for generating high quality and realistic human motions.

\begin{table}[t]
    \centering
    \small
        \caption{Ablation study on varying the number of Bi-Temporal Mamba blocks ($N$) and the number of layers ($L$), evaluated on HumanML3D.}
        \vspace{-1.5mm}
    \resizebox{0.90\columnwidth}{!}{
    \begin{tabular}{l@{\hspace{0.4cm}}c@{\hspace{0.4cm}}c@{\hspace{0.4cm}}c@{\hspace{0.4cm}}c}
    \toprule
        N & L &R Precision (Top 3)$\uparrow$ & FID$\downarrow$ & MM Dist$\downarrow$ \\
    \midrule
    2 & 2 
      & $0.813^{\pm .002}$
      & $0.069^{\pm .002}$
      & $2.907^{\pm .005}$ \\
    \textbf{3} & \textbf{2}
      & ${0.813}^{\pm .002}$
      & $\textbf{0.039}^{\pm .002}$
      & $2.884^{\pm .008}$ \\
    4 & 2
      & $\textbf{0.817}^{\pm .002}$
      & $0.074^{\pm .002}$
      & $\textbf{2.872}^{\pm .007}$ \\
    2 & 3
      & $0.814^{\pm .003}$
      & $0.062^{\pm .003}$
      & $2.893^{\pm .007}$ \\
    3 & 3
      & $0.810^{\pm .002}$
      & $0.054^{\pm .003}$
      & $2.914^{\pm .008}$ \\
    4 & 3
      & $0.816^{\pm .002}$
      & $0.052^{\pm .002}$
      & $2.873^{\pm .008}$ \\
    \bottomrule
    \end{tabular}
    }
    \label{tab:Layers}
    \vspace{-2mm}
\end{table}

\begin{table}[t]
\centering
\small
\caption{Ablation study on key components, evaluated on HumanML3D. 
We use \textbf{Part-b} and \textbf{Whole-b} to denote the part-based learning and 
whole-based learning in \emph{Dual-Spatial Mamba}, respectively.}
\vspace{-1.5mm}
\label{tab:comparison_ablation}
 \resizebox{1\columnwidth}{!}{
\begin{tabular}{c@{\hspace{0.4cm}}c@{\hspace{0.4cm}}c@{\hspace{0.4cm}}c}
\toprule
& R-Top3$\uparrow$ & FID$\downarrow$ & MM Dist$\downarrow$\\ 
\midrule
w/o DSFM & $ \textbf{0.817}^{\pm .002}$ & $0.096^{\pm .003}$ & $2.889^{\pm .010}$ \\
w/o Part-b            & $0.816^{\pm .002}$ & $0.066^{\pm .003}$ & $\textbf{2.869}^{\pm .008}$ \\
w/o Whole-b           & $0.812^{\pm .002}$ & $0.065^{\pm .003}$ & $2.901^{\pm .009}$ \\
w/o Part-b $\&$ Whole-b & $0.815^{\pm .002}$ & $0.070^{\pm .003}$ & $2.874^{\pm .008}$ \\
HiSTF Mamba (Ours)    &   ${0.813}^{\pm .002}$
      & $\textbf{0.039}^{\pm .002}$
      & $2.884^{\pm .008}$ \\
\bottomrule
\end{tabular}}
    \label{core module}
\vspace{-4mm}
\end{table}
\vspace{-1mm}

\subsection{Implementation Details}
We implement HiSTF Mamba using the AdamW optimizer, maintaining a fixed learning rate of $1 \times 10^{-4}$. The training process spans 350 epochs with a batch size of 128. Regarding the diffusion configuration, we establish the number of training steps at 1000, while employing the UniPC~\cite{zhao2023unipc} scheduler with 20 steps for sampling to ensure high quality generation. The entire training phase requires approximately 18 hours on two NVIDIA A100 GPUs using bfloat16 mixed precision to enhance efficiency. For inference, the model achieves an average latency of $0.233$ seconds on a single NVIDIA A100 GPU. Furthermore, our architecture remains lightweight with a trainable parameter count of $23.95 M$.

\section{Conclusion}
\label{conclusion}

In this work, we presented HiSTF Mamba, a linear complexity framework for efficient text driven motion generation. By synergizing Dual Spatial Mamba with the Dynamic Spatiotemporal Fusion Module, our model captures fine joint dynamics and global coordination while eliminating feature redundancy. Beyond demonstrating superior experimental performance, this work offers the community a promising direction for developing lightweight and high fidelity generative models.

\bibliographystyle{IEEEbib}
\bibliography{icme2026references}

\end{document}